\documentclass[referee]{sn-jnl}

\usepackage[utf8x]{inputenc}
\usepackage{bbold} 

\usepackage{amsmath,amsthm,amssymb}
\usepackage{graphicx}

\newtheorem{defn}{Definition}

\newtheorem{corollary}{Corollary}
\newtheorem{theorem}{Theorem}

\newcommand{\sS}{\mathcal S}
\newcommand{\sA}{\mathcal A}

\jyear{2021}%

\begin{document}

\title[Transfer in Reinforcement Learning via Regret Bounds for Learning Agents]{Quantification of Transfer in Reinforcement Learning via Regret Bounds for Learning Agents}
\author[1]{\fnm{Adrienne} \sur{Tuynman}}\email{adrienne.tuynman@inria.fr}

\author*[2]{\fnm{Ronald} \sur{Ortner}}
\email{rortner@unileoben.ac.at}

\affil[1]{\orgname{Scool team, Univ. Lille, Inria, CNRS, Centrale Lille, UMR 9189 CRIStAL, F-59000 Lille}, \country{France}}

\affil[2]{\orgdiv{Lehrstuhl für Informationstechnologie}, \orgname{Montanuniversität Leoben}, \country{Austria}}


\abstract{
We present an approach for the quantification of the usefulness of transfer in reinforcement learning via regret bounds for a multi-agent setting. Considering a number of $\aleph$ agents operating in the same Markov decision process, however possibly with different reward functions, we consider the regret each agent suffers with respect to an optimal policy maximizing its average reward. We show that when the agents share their observations the mutual regret of all agents is smaller by a factor of $\sqrt{\aleph}$ compared to the case when each agent has to rely on the information collected by itself. This result demonstrates how considering the regret in multi-agent settings can provide theoretical bounds on the benefit of sharing observations in transfer learning. 
}
       

\maketitle

\section{Introduction}

Transferring relevant knowledge to apply it to a variety of similar environments is not a new concept: in psychology and cognitive sciences, it is well known that humans learn better when they are already familiar with a similar task. This concept helped form \textit{transfer learning}, a wide and important area of machine learning with a lot of empirical studies, but few theoretical guarantees available, see e.g.\ \cite{pan} for a survey. 
This holds true even more in the specific area of transfer in reinforcement learning (RL), which presents some additional challenges. In an RL problem the learner when set a task has to collect information in the underlying domain by itself with the aim of finding a successful policy fulfilling the task. The question of transfer is to what extent the collected information and the learned policy can be useful for a different task in the same or a similar domain. One of the problems is that in general the different task may demand a completely different sampling strategy from the learner so that observations for a different problem setting may turn out to be not very useful. This makes the question of transfer already difficult for a concrete problem setting and it is even harder to come up with any theoretical results that generalize beyond.  

Accordingly, the two survey papers of Taylor \& Stone \cite{taylor-transfer} and Lazaric \cite{lazaric-transfer} are mainly dedicated to establishing a taxonomy of various settings as well as used performance criteria investigated in the literature. One particular problem is the quantification of the benefit of transfer in RL, for which there are hardly any general results. While Taylor \& Stone \cite{taylor-transfer} present different criteria for the evaluation of transfer learning methods, these are rather discussed with respect to application to empirical studies.

Among the considered criteria are the \textit{total reward}, that is, the total amount of accumulated reward, and the \textit{transfer ratio}, that is, the ratio of total reward accumulated with transfer over the total reward accumulated without transfer. This latter criterion is the closest in the literature to what we suggest in this paper, that is, to consider the \textit{regret} in order to measure the utility of re-using samples. The regret compares the accumulated rewards of a learning agent to that of an optimal policy. In what follows we aim to show that the regret of an agent that learns to solve some task becomes smaller when it uses samples collected during learning some other task.

\subsection{Mutual Regret in a Multi-Agent Setting}\label{sec:marl}

Our results are derived in a more general multi-agent setting with several agents simultaneously learning different tasks in the same environment, represented by a Markov decision process (MDP) with a transition function common to all agents but specific reward functions for each single agent. The agents share their observations and we are interested in quantifying the usefulness of the shared information. For this we propose to compare the \textit{mutual regret} over all agents to the regret in the standard (single-agent) RL setting when no additional information is available. The respective improvement in regret can be interpreted in terms of mutual transfer between the learning agents. 

\subsection{Related Work}\label{sec:marl}

While Taylor \& Stone \cite{taylor-transfer} mention multi-agent settings as an interesting application area for transfer (cf.\ the related survey \cite{velo}), there is also a particular survey of da Silva and Costa \cite{silva}, which discusses transfer for multi-agent RL. In general, whereas we will concentrate on MDPs in which all agents share a common transition function that only depends on the chosen action by each individual agent, work on multi-agent RL (MARL) settings often rather considers the more complex \textit{stochastic game} scenario where transitions depend on the joint actions chosen by the agents. This also gives rise to different goals ranging from Nash equilibria and cooperative learning to adversarial settings in which one is interested in algorithms for a single self-interested agent. Applications for such settings can be e.g.\ found in game playing, fleet balance, autonomous driving, or robotics, see \cite{MARL_overview,MARL_deep,MARLPPO}. Research on transfer learning in MARL settings often  provides general frameworks and architectures~\cite{MATLYang,MATLZhou}, also for cross-task transfer \cite{shi-lateraltransfer-marl}, however there are hardly any general results available.

Our approach rather fits the inter-agent transfer learning setting discussed in  the survey \cite{dasi-interagent-transfer}: Rather than having a single agent that transfers knowledge of its previous tasks, we have several agents simultaneously learning from each other. In our setting they are solving different tasks, however do this in the same environment. 

Although only loosely related to our research, we briefly point out some work on transfer in multi-agent \textit{deep} RL. In general, transfer in (single-agent) deep RL has its own specific challenges, the survey of~\cite{zhu-deeptransfer} provides a brief overview of different topics and methods. A respective survey on multi-agent deep RL is also available~\cite{du-madeeprl}, a more recent review focuses on the cooperative setting~\cite{oro-coop-madeeprl}. As theoretical results are already scarce for deep RL, we are not aware of any respective guarantees in the more general multi-agent settings.

More importantly, the setting of multiple agents collaborating on a single problem has been dealt with in the literature under various different names.
In \textit{fleet learning}, the agents do not interact with each other, but instead act at the same time in similar environments. This is used in practice for various uniform machines with slight differences in manufacturing, such as wind farms~\cite{verstraetenFleet}. Unlike that, in our setting, the agents share the environment but may have completely different tasks.

In \textit{distributed learning} one aims to scale up learning by using multiple parallel agents on the same task, see e.g.~\cite{DistributedImpala}.
This idea has also been used for more difficult learning settings such as POMDPs~\cite{distrRL-POMDP}. We generally consider the agents having different tasks, however also show some improvements in theory and experiments when all agents try to solve the same task. Moreover, we are  not aware of any other work that considers regret in a distributed RL setting.  

Another related concept is \textit{federated learning}, which however puts more emphasis on different topics such as privacy and limiting the amount of data that is shared, see \cite{FedRLsurvey,Federated2022}, which is something that is not relevant in our setting.

\subsection{Contributions and Outline}

We consider a multi-agent setting in which all agents have the same state and action space. The transferred knowledge is the experience of each single agent (also called \textit{instance transfer}), which is shared with all other agents. It is assumed that each agent learns a different task, however the agents act in the same environment. That is, the transition probabilities are the same, while each agent as its own reward function. 

For this setting, we propose an optimistic learning algorithm that is an adaptation of a single-agent RL algorithm to the multi-agent case and for which we can show regret bounds that are able to quantify the benefit of transfer in the considered setting. Implicitly our multi-agent algorithm also contributes to an issue pointed out by Lazaric \cite{lazaric-transfer} who stated that ``\textit{the problem of how the exploration on one task should be adapted on the basis of the knowledge of previous related tasks is a problem which received little attention so far.}''

As already mentioned we are interested in regret bounds that measure for any time step $T$ the difference of the accumulated reward of an agent to the total reward an optimal policy (for the respective agent) would have achieved. We will see that while it is difficult to say how much benefit the experience of a single agent contributes to learning another task by a different agent, when considering the \textit{mutual regret} over all agents the worst case regret bounds are smaller by a factor of $\sqrt{\aleph}$ for a total of $\aleph$ agents. 

These are the first theoretical regret bounds for reinforcement learning with multiple simultaneously learning agents in MDPs we are aware of. Our theoretical findings are complemented by experiments on two toy domains, which confirm that the regret in the joint learning setting per agent is smaller than in the standard RL case. From the perspective of transfer learning, these results demonstrate the mutual utility of samples collected by agents learning different tasks in the same environment both practically and theoretically.
\bigskip

The paper is organized as follows: We start with some preliminaries about MDPs in Section~\ref{sec:mdps}. Section~\ref{sec:transfer}
defines the precise setting of transfer learning we are interested in and introduces the notion of \textit{mutual regret} in a multi-agent setting. Then we present our algorithm, for which we also provide respective performance bounds in the main Section~\ref{sec:alg}, followed by a detailed discussion about the interpretation of our results from the perspective of transfer learning in Section~\ref{sec:disc}. Section~\ref{sec:exp} presents the results of some complementary experiments. The paper is concluded by Section \ref{sec:conclusion} considering a few open questions and possible directions of future work.

%
%

\section{MDP Preliminaries}\label{sec:mdps}

Before we introduce the transfer learning setting we consider, we start with some preliminaries on  Markov decision processes (MDPs), which will be used to represent single learning tasks.

\begin{defn}
A Markov decision process $M=(\sS,\sA,p,r,s_1)$ consists of a set of states $\sS$ of finite size $S$
and a set of actions $\sA$ of finite size $A$ available in each of the states.\footnote{It is straightforward 
but notationally a bit awkward to generalize our setting as well as the results to the case
where different actions are available in each of the states.} When choosing an action 
$a \in \sA$ in state $s \in \sS$, the agent obtains a stochastic reward with mean $r(s,a)$ and moves to a new state~$s'$ that is determined by the transition probabilities $p(s'\!\mid\! s,a)$. The learning agent starts in an initial state $s_1$.
\end{defn}

A \textit{policy} in an MDP defines which actions are taken when acting in the MDP. In the following, we will only consider \textit{stationary} policies $\pi:\sS\to \sA$ that pick in each state $s$ a fixed action $\pi(s)$. For the average reward criterion we will consider, it is sufficient to consider stationary policies (cf.~Section~\ref{sec:goal} below).

\subsection{Assumptions and the Diameter}

To allow the agent to learn, we make further assumptions about the underlying MDP. First, with unbounded rewards it is difficult to learn (as it is easy to miss out a single arbitrarily large reward at some step), so we assume bounded rewards. Renormalizing, in the following we make the assumption that all rewards are contained in the unit interval $[0,1]$.

Further, we assume that it is always possible to recover when taking a wrong action once. This is possible if the MDP is \textit{communicating}, i.e., any state is reachable from any other state with positive probability when selecting a suitable policy. That is, given a communicating MDP and two states $s,s'$ there is always a policy $\pi$ such that the expected number of steps $T_\pi(s,s')$ that an agent executing policy $\pi$ needs to reach state $s'$ when starting in state~$s$ is finite. 
Next, we introduce the diameter of the MDP, a related parameter we will use in the following that measures the maximal distance between any two states in a communicating MDP.

\begin{defn}
 The diameter $D$ in a communicating MDP $M$ is defined as 
\[
   D(M) := \max_{s,s'\in\sS} \min_\pi  T_\pi(s,s').
\]

\end{defn}

\subsection{The Learning Goal and the Notion of Regret}\label{sec:goal}

The goal of the learning agent is to maximize its accumulated reward after any $T$ steps. For the sake of simplicity, in the following we will however consider the \textit{average reward}
\[
   \lim_{T\to \infty} \frac{1}{T} \sum_{t=1}^T  r(s_t,a_t),
\]
where $s_t$ and $a_t$ are state and action at step $t$, respectively. Unlike the accumulated $T$-step reward, the average reward is known to be maximized by a stationary policy $\pi^*$~\cite{puterman}. The optimal average reward $\rho^*$ is a good proxy for the (expected) optimal $T$-step reward, as $T \rho^*$ differs from the latter at most by a term that is of order $D$, as noted by \cite{jaorau}. Accordingly, we measure the performance of the learning agent by considering its \textit{regret} after any $T$ times steps defined as
\[
     R_T := T \rho^*  - \sum_{t=1}^T r_t,
\]
where $r_t$ is the reward obtained by the agent at step~$t$.

Literature provides bounds on the regret for various algorithms. One of the first algorithms for which regret analysis has been provided is UCRL2 \cite{jaorau}, an algorithm that implements the idea of using optimism in the face of uncertainty and that inspired several variations based on the same idea~\cite{KLUCRL,ronan4,ucrl3}. For UCRL2, a regret bound of order $\tilde{O}({DS\sqrt{AT}})$ has been given, complemented by lower bound of $\Omega(\sqrt{DSAT})$ showing that the upper bound is close to optimal. In the meantime, the gap between upper and lower bound on the regret has been closed in \cite{boone} for another optimistic algorithm in the spirit of UCRL2. We will employ UCRL2 as a base for our multi-agent learning algorithm in Section~\ref{sec:alg}.

\section{Transfer Learning}\label{sec:transfer}

In this section, we first suggest how regret can be used to quantify the benefit of transfer in RL in a sequential setting, before introducing the related notion of \textit{mutual regret} in a multi-agent setting.

\subsection{Sequential Transfer Learning}\label{sec:vanilla}

We are interested in the situation when a learning agent is presented two or more tasks in a row, each represented by an MDP on the same state-action space. That is, we have a finite sequence of MDPs $M_1, M_2, \ldots, M_\aleph$ and the learner operates in each MDP $M_\alpha$ for $T_\alpha$ steps, before the task changes to the next MDP $M_{\alpha+1}$. (Unlike the tasks themselves, these change points are known to the learner.) 

The transfer learning paradigm we suggest is as follows. Consider the regret of a learning algorithm that treats the tasks independently of each other and compare it to the regret of a modified algorithm that uses in each task $M_\alpha$ the information gathered in previous tasks $M_1,M_2,\ldots,M_{\alpha-1}$. The difference in regret obviously is a quantification of the benefit of transfer. 

Obtaining bounds for transfer learning in a given task sequence seems to be hard however, cf.\ the open questions at the end of the paper. What we managed to achieve are results for a related measure which we call \textit{mutual regret} in a multi-agent setting, which we introduce in the following section.

\subsection{Mutual Transfer and Regret in a Multi-Agent Setting}
While transfer as defined in the previous section depends on a fixed task sequence $M_1, M_2, \ldots, M_\aleph$, we rather take a look at the joint transfer over all permutations of these tasks. More precisely, we consider the following multi-agent setting, which generalizes the introduced RL setting to more than one agent. There are $\aleph$ learning agents $\alpha=1,2,\ldots,\aleph$ which simultaneously act in their own MDP $M_\alpha$. We assume however that all agents share their experiences. That is, each agent has immediate access to all the observations made by each other agent~$\alpha$, in particular its history $(s^\alpha_t,a^\alpha_t,r^\alpha_t)_{t\geq 0}$ specifying the state $s^\alpha_t$ visited at step~$t$, the action $a^\alpha_t$ chosen and the reward $r^\alpha_t$ obtained. (As each learning agent aims to maximize its own rewards, knowledge of the other agents' rewards is only of use in the special case when all agents have the same reward function, cf.~Theorem~\ref{thm2} below.) 

In terms of the transfer learning setting introduced before, this means that each agent learns its own task $M_\alpha$, however having at any step $t$ access to the samples collected by the other agents in their learning tasks up to the same step $t$. Unlike in the sequential setting where we assume that the learner has \textit{all} the information on the previous tasks, this means a more limited access and accordingly a harder task.

In what follows we consider that all agents act in the same environment, thus having not only the same state-action space but also common transition probabilities. However, each agent $\alpha$ is assumed to have a different task, represented by a reward function $r^\alpha(s,a)$ specific to the agent $\alpha$. 

Then for each agent $\alpha$ we can define its \textit{individual regret} as 
\[
     R^\alpha_T := T \rho^{*,\alpha}  - \sum_{t=1}^T r^\alpha_t,
\]
where $\rho^{*,\alpha}$ is the optimal average reward under the reward function $r^\alpha$. The individual regret can be considered as a proxy for the regret in the sequential transfer learning setting for which it seems difficult to obtain meaningful results. In the following we will instead consider the \textit{mutual regret} defined as the sum of the individual regret over all agents, i.e.,
\[
   R^{\rm m}_T :=  \sum_{\alpha} R^\alpha_T .
\]
We stick to the idea we have suggested for the sequential transfer setting: Our aim is to compare the mutual regret averaged over all agents to the regret when learning one of the tasks $M_\alpha$ in isolation (i.e., without transfer). The following section shows that one can indeed obtain respective improved regret bounds for the multi-agent case, showing that one can quantify the benefit of transfer via bounds on the mutual regret.

\section{Regret Bounds for Multi-Agent Learning} \label{sec:alg}
In this section, we propose an algorithm for the introduced multi-agent setting and present upper bounds on the mutual regret of all learning agents.

\subsection{UCRL2 with Shared Information}
For our approach we propose Multi-agent-UCRL (shown in Algorithm 1), a variant of the reinforcement learning algorithm UCRL2 \cite{jaorau} that is able to make use of shared information. That is, each agent will select its actions according to UCRL2 (briefly introduced below) however using not only information collected by itself but also by all other agents.

\begin{algorithm}[!t]
\caption{Multi-agent-UCRL with shared information}\label{alg}
\begin{algorithmic}[1]
\item \textbf{Input:} State space $\sS$, action space~$\sA$, number of agents $\aleph$, confidence parameter $\delta$.
\item \textbf{Initialization:} Observe the initial state $s_1$. \\ 
In the following, let $t$ be the current time step.
\item \FOR {episodes $k = 1,2,\ldots$:} 
\item Compute estimates $\hat{r}^\alpha_k(s,a)$ for rewards of each agent $\alpha$ and estimates $\hat{p}_k(\cdot\mid s,a)$ for transition probabilities (common to all agents). \\
\textbf{Compute policy $\tilde{\pi}_k^\alpha$ for each agent $\alpha$:}
\item Let $\mathcal{M}_k^\alpha$ be the set of plausible MDPs $\tilde{M}$ with rewards $\tilde{r}(s,a)$ and transition probabilities $\tilde{p}(\cdot\mid s,a)$ close to the computed estimates 
\begin{align*}
 \mid \tilde{r}(s,a) - \hat{r}_k^\alpha(s,a)\mid  &\leq\, {\rm conf}_r^\alpha(s, a, S, A, \aleph, \delta, t),  \\
 \big\| \tilde{p}(\cdot\mid s,a) - \hat{p}_k(\cdot\mid s,a) \big\|_1 &\leq\, {\rm conf}_p(s, a, S, A, \aleph, \delta, t). 
\end{align*}
\item For each agent $\alpha$ compute an MDP $\tilde{M}_k^\alpha$ in $\mathcal{M}_k^\alpha$ with
\begin{equation*}
 \rho^*(\tilde{M}^\alpha_k) = \max \{\rho^*(M) \mid  M \in \mathcal{M}_k^\alpha \}.
\end{equation*}
Further, let $\tilde{\pi}^\alpha_k$ be the respective optimal policy in $\tilde{M}_k^\alpha$.\footnotemark
\item \textbf{Let each agent $\alpha$ execute policy} $\tilde{\pi}^\alpha_k$:
\item  Let each agent $\alpha$ \\
  $\rhd$ choose action $a^\alpha_t = \tilde{\pi}_k^\alpha(s^\alpha_t)$, \\
   $\rhd$ obtain reward $r^\alpha_t$, and \\
  $\rhd$ observe state $s^\alpha_{t+1}$.
\item  If the number of visits of some agent in some state-action pair has been doubled since the start of the episode, terminate current episode and start new one.
\end{algorithmic}
\end{algorithm}

UCRL2 is a model-based algorithm that implements the idea of optimism in the face of uncertainty. 
\footnotetext{\label{fn:evi}The computation of $\tilde{\pi}^\alpha_k$ and $\tilde{M}_k^\alpha$ can be done using \textit{extended value iteration} as introduced in~\cite{jaorau}. For the proof of the regret bound we will assume that extended value iteration is employed with accuracy $1/\sqrt{t}$.}
Basically, UCRL2 uses the collected rewards as well as the transition counts for each state-action pair $(s,a)$ in order to compute estimates $\hat{r}$, $\hat{p}$ for rewards and transition probabilities, respectively (cf.\ line 5 of Algorithm 1). While the original algorithm only considers a single agent, in the shared information setting the estimates are computed from the collected observations of all agents. Based on the estimates and respective confidence intervals a set $\mathcal M^\alpha$ of plausible MDPs is computed for each agent~$\alpha$ (line~7). These confidence intervals are chosen so that on the one hand they shrink with an increasing number of observations but on the other hand always contain the true values of the underlying MDP with high probability. Different choices for these confidence intervals have been suggested in the literature. Here we stick to the confidence intervals originally used for UCRL2 \cite{jaorau}, however point out that one can e.g.\ use tighter Bernstein style concentration inequalities such as suggested in \cite{ucrl3}.  Instead we use simpler Hoeffding concentration inequalities and define the confidence intervals used by agent $\alpha$ at step~$t$ as
\begin{eqnarray}
  {\rm conf}_r^\alpha(s, a, S, A, \aleph, \delta, t) &:=&  \sqrt{\frac{7\log\left({\frac{2SA\aleph t}{\delta}}\right)}{2\max{\{1,N_t^\alpha (s,a)\}}}},  \label{eq:cr} \\[2\jot]
  {\rm conf}_p(s, a, S, A, \aleph, \delta, t) &:=&  \sqrt{\frac{14S\log\left({\frac{2A t}{\delta}}\right)}{\max{\{1,N_t(s,a)\}}}}, \label{eq:cp}
\end{eqnarray}

\noindent
where $N_t^\alpha(s,a)$ is the number of samples taken by agent $\alpha$ in $(s,a)$, and $N_t(s,a):=\sum_{\alpha=1}^\aleph N_t^\alpha (s,a)$ is the total number of samples taken by agents in $(s,a)$. As all agents have the same transition function, but the reward functions are individual, the confidence intervals for rewards depend only on the number of visits in $(s,a)$ by the agent~$\alpha$, while the confidence intervals for the transition probabilities depend on the number of samples shared by all agents. Moreover, in order to account for the $S$ different outcomes of a sample from the transition probability distribution, the confidence intervals in \eqref{eq:cr2} have an additional factor of $\sqrt{S}$, cf.\ the bounds provided by \cite{weissman03}, which are used in the regret analysis for UCRL2 (and implicitly also for our results).

Each agent~$\alpha$ then optimistically picks an MDP $\tilde{M}^\alpha\in\mathcal M^\alpha$ and a policy~$\tilde{\pi}^\alpha$ that maximize the average reward (line~8). This policy is played until the number of visits in some state-action pair has been doubled for some agent, when a new policy is computed (lines~10--14). The phases between computation and execution of a new policy are called \textit{episodes}.

We note that while the algorithm is described as a central algorithm controlling all $\aleph$~agents, it could also be rewritten as a local algorithm for each single agent having access to the observations of the other agents.

\subsection{Upper Bound on the Mutual Regret}
The following upper bound on the mutual regret of all agents using Multi-Agent-UCRL holds. 
\begin{theorem}\label{thm}
With probability $1-\delta$, after any $T$ steps the mutual regret of agents controlled by Multi-agent-UCRL with confidence intervals \eqref{eq:cr} and \eqref{eq:cp} is upper bounded by 
\[
   R^{\rm m}_T  \,  \leq \, 15 \left( D\sqrt{S}+\sqrt{\aleph}\right) \sqrt{SA\aleph T\log\left( \tfrac{8A\aleph T}{\delta}\right)} .
\]
\end{theorem}

\begin{corollary}\label{cor}
If $\aleph<D\sqrt{S}$, the average mutual regret per agent after $T$ steps is $\tilde{O}\left(\frac{DS\sqrt{AT}}{\sqrt{\aleph}}\right)$.
\end{corollary}

Corollary \ref{cor} shows that averaged over the agents the upper bound on the (mutual) regret improves over the respective bound on the individual regret of a single agent. As already mentioned, the latter is of order $\tilde{O}({DS\sqrt{AT}})$ in the standard reinforcement learning setting for the UCRL2 algorithm, see~\cite{jaorau}.  Appendix~\ref{app} provides a sketch of the proof of Theorem~\ref{thm}, which is based on the analysis of \cite{jaorau}. The main reason for the improvement being possible is that learning the transition function in general is harder in the sense that the respective error has a larger influence on the regret. As several agents are able to learn the common transition function faster, this results in improved bounds.

Further improvement is possible if the agents also have a joint reward function. While this setting is not relevant in the context of transfer learning, it still may be interesting in the context of distributed learning,
when all agents share the same MDP. In this case it makes sense to use common confidence intervals also for the rewards that now do not depend on the individual visits $N^\alpha(s,a)$ of an agent $\alpha$ but on the number of total visits $N(s,a)$ of all agents. That is, confidence intervals 
\begin{eqnarray}
  {\rm conf}_r^\alpha(s, a, S, A, \aleph, \delta, t) &:=&  \sqrt{\frac{7\log\left({\frac{2SA t}{\delta}}\right)}{2\max{\{1,N_t(s,a)\}}}}.  \label{eq:cr2} 
\end{eqnarray}
can be used for the rewards.
Accordingly, also  the episode termination criterion should be changed to depend not on the visits of an individual agent but on the total number of visits. That is, an episode will be terminated if the total number of visits of all agents has been doubled since the start of the episode. With these modifications, one can obtain regret bounds of order $\tilde{O}(DS\sqrt{A\aleph T})$.

\begin{theorem}\label{thm2}
In a setting where all agents share the same reward function, the mutual regret of Multi-agent-UCRL (with confidence intervals \eqref{eq:cr2} and using a modified episode termination criterion) with probability $1-\delta$ after any $T$ steps is upper bounded by 
\[
   \sum_{\alpha=1}^\aleph R^\alpha_T  \,  \leq \, 34  D S \sqrt{A\aleph T \log\left( \tfrac{8A\aleph T}{\delta}\right)} .
\]
\end{theorem}

We highlight the necessary modifications in the proof of Theorem~\ref{thm} to obtain Theorem~\ref{thm2} in Appendix \ref{app2}.
By Theorem \ref{thm2}, Corollary~\ref{cor} holds without any condition on the number of agents in the setting of shared rewards.
As far as we are aware of, these are the first results on the regret in a distributed RL setting.

\section{Discussion}\label{sec:disc}

The results presented in Theorems \ref{thm} and \ref{thm2} are not only the first regret bounds for simultaneously learning agents in a MARL setting. They also provide a quantification for the utility of transfer along the lines sketched in Section~\ref{sec:transfer}. Comparing the average mutual regret of an agent (Corollary \ref{cor}) to the bounds of the single agent setting we can see an improvement due to the use of the samples collected by the other agents learning their tasks. While the bounds are derived for mutual transfer in a multi-agent setting, they also provide some information about the sequential transfer in standard (i.e., single agent) RL. That is, the mutual regret in the multi-agent setting can be considered to be a proxy for the average regret when learning a task $M_\alpha$ with samples from tasks $M_1,\ldots,M_{\alpha-1},M_{\alpha+1},\ldots,M_\aleph$ available, where the average is taken over these $\aleph$ learning tasks. As already indicated, learning in the multi-agent setting is harder as less information is available: In the sequential transfer setting all samples of previous tasks are available, while in the multi-agent setting each agent only has the samples of other agents up to the current time step at its disposal.  

Notably, while our results demonstrate a collective benefit over all tasks/agents, it seems difficult to obtain any results for the utility of samples collected when learning a specific task $M_\alpha$ for learning a different particular task $M_{\alpha'}$. Thus, while on the one hand individual transfer seems to be difficult to quantify, on the other hand a general mutual transfer is provable. 


In the analysis, the role of a single agent is similar to that of a single state-action pair in UCRL2. That is, while it seems difficult to say how much the learning process focuses on the states and actions relevant for the task to learn, the regret analysis uses that one can at least bound the sum of visits over all state-action pairs, cf.\ eq.~\eqref{eqr3a} in the analysis.



For a start the choice of an RL standard algorithm like UCRL2 looks quite natural. More generally, we think that our approach can be easily adapted to other algorithms for which similar theoretical guarantees are available and hence can serve as a blueprint for further investigations on transfer in RL. It would be particularly interesting to see whether in the presented framework it turns out that some RL algorithms make better use of additional samples coming from other tasks than other algorithms.

\section{Numerical experiments}\label{sec:exp}
In this section, we report some experimental results on the two domains \textit{RiverSwim} and \textit{SixArms} taken from \cite{stli3}. These results complement our theoretical findings and show that the improvement in the regret bounds (which are worst case guarantees) are not just an artifact of our theoretical analysis and that agents sharing their information indeed suffer smaller regret.


%
\subsection{Setting and Implementation}
For the implementation of our multi-agent algorithm we used for each agent  the same parameter for $\delta$ set to\footnote{Choosing a rather large $\delta$ is equivalent to using smaller confidence intervals, which usually results in faster convergence of UCRL2 to an optimal policy. This is because the original confidence intervals are chosen quite conservatively in order to be able to show the bounds on the regret.} $\delta=0.5$. 
We investigated the same domains having a different total number of agents $\aleph$ in order to demonstrate that the advantage of joint learning increases with the number of agents.

The first setting is the \textit{SixArms} environment \cite{stli3} with seven states and six actions. The learner starts in a central state from which six reward states are reachable  when choosing the right action. While the central state can be reached deterministically, the reward states are differently hard to reach with transition probabilities reaching from 0.01 to 1. That way, for some states  the learner has to explore a lot in order to reach them. When picking the right action each of the six reward states $s_i$ gives a different mean reward of $\frac{i}{5}$ with $i=0,1,\ldots,5$, and the agent needs to identify the state and action that provide the highest reward. In our experiment, the various agents had the different rewards randomly shuffled among the reward states, so that the optimal state was not fixed but dependent on the agent. That way, agents are expected to spend more time exploring their own promising states, while doing less useful exploration for other agents.

The second setting we considered is the 6-state \textit{RiverSwim} problem also taken from \cite{stli3}, in which agents have to choose a direction that leads with some probability to one of two possible neighbor states. The agents have to learn to abandon safety (and a small reward of 0.05) of the initial state and go through a sequence of no-reward states in order to reach a high reward state (on the other side of the river) giving a maximal reward of 1. Here all agents shared the same reward function. While this setting is not interesting from the perspective of transfer, we performed experiments on it for the confirmation of the results of Theorem \ref{thm2}.

\subsection{Results}

For the results we report the mutual regret over time averaged over the number of agents (and all experiments), as well as the first and third quartiles of that mean. In a separate graph, we display the regret achieved by each single agent in one arbitrary run of the multi-agent UCRL algorithm.

\begin{figure}[!ht]
	\centering
\includegraphics[width=0.49\linewidth]{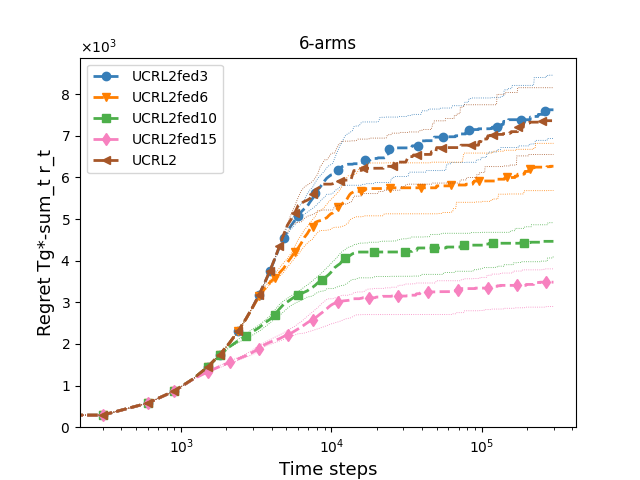}
\includegraphics[width=0.49\linewidth]{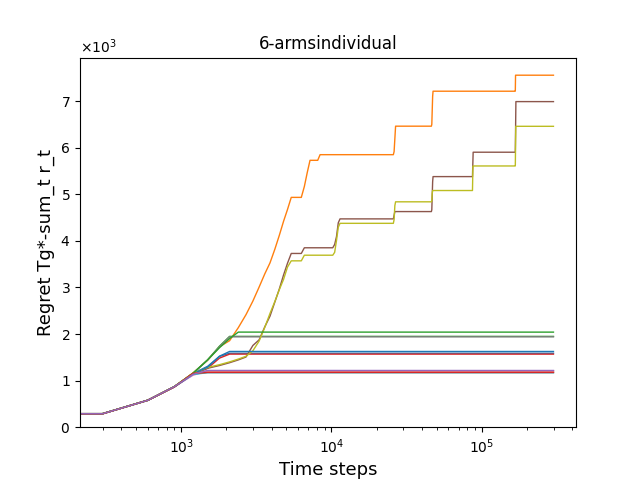}
\caption{An experiment in the \textit{SixArm} environment, with (left) the average mutual regret for $\aleph=1$ (UCRL2) and $\aleph=3,6,10,15$ over 64 runs, and (right) the regret of each of the 15 agents of Multi-agent-UCRL with shared information in one run.}
\label{fig:exp_6arms}
\end{figure}

Figure~\ref{fig:exp_6arms} shows the results for the \textit{SixArms} environment. When there are three agents for the six states to explore, the regret is similar to that of UCRL2 in the ordinary (i.e., single agent) RL setting. However, with an increasing number of agents the regret drops considerably. Although the agents have different target states, states with low reward need less exploration as observations for estimating the transition probabilities are in particular provided by other agents for which this state is more promising. Note that while with 15 agents there will be agents sharing the same optimal arm, this information in itself cannot be exploited by any of the agents since the identity of these agents is unknown. When taking a look at a single run, it can be noticed that the regret of the single agents is quite differently distributed (unlike in the \textit{RiverSwim} domain discussed below). This is because the agents have different tasks some of which are harder (when the state with the highest reward is difficult to reach), while others are quite easy.

\begin{figure}[!ht]
	\centering
\includegraphics[width=0.49\linewidth]{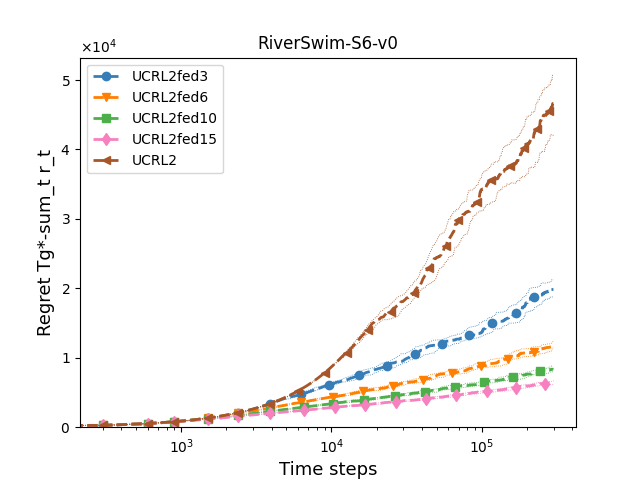}
\includegraphics[width=0.49\linewidth]{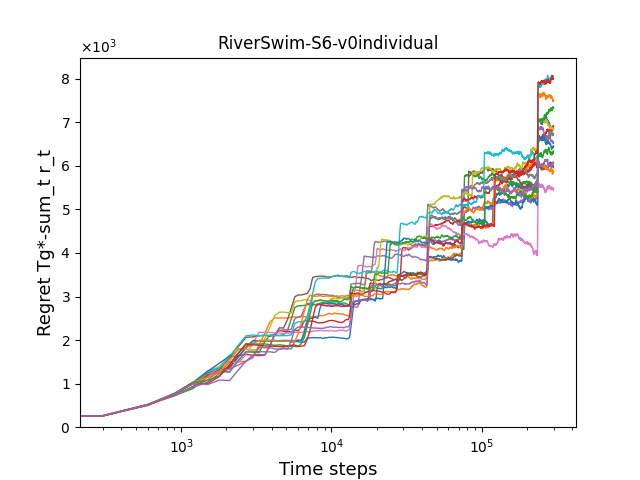}
\caption{An experiment in the \textit{RiverSwim} environment, with (left) the average mutual regret for $\aleph=1$ (UCRL2) and $\aleph=3,6,10,15$ over 64 runs, and (right) the regret of each of the 15 agents in one run of Multi-agent-UCRL with shared information.}
\label{fig:exp_rs}
\end{figure}

The results for \textit{RiverSwim} are displayed in Figure~\ref{fig:exp_rs}. We see that the average mutual regret is below that of the single agent setting and also decreases with the number of agents learning together. Moreover, when studying a single arbitrary run, we notice that the regret seems to be quite fairly divided among the various agents, though we only have theoretical guarantees for the mutual regret over all agents.

Note that the experimental results give much lower regret than the upper bounds of Theorems \ref{thm} and \ref{thm2}. This is due to the conservative nature of the confidence intervals we use. Usually, the empirical performance of UCRL2 and similar algorithms can be improved by further reducing the size of the confidence intervals, however at the cost of losing the theoretical guarantees.

Overall, the results confirm that the regret is reduced when sharing information. This holds not only when the task is the same as in the \textit{RiverSwim} example, but also when the agents have different goal states and only the transition probabilities are shared.

\section{Open Questions and Future Work}\label{sec:conclusion}

Our aim was to provide theoretical results for the benefit of transfer in reinforcement learning. The multi-agent setting we have considered allows to illustrate that joint use of samples collected when simultaneously learning different tasks is provably useful overall. This has been formalized by our notion of \textit{mutual regret} which shrinks with the number of agents corresponding to the number of tasks in our transfer learning setting. 

While we think that our approach is of general interest, as it leads to otherwise hardly available theoretical guarantees for transfer in reinforcement learning, the results we have achieved are only a starting point for further research and moreover leave a few open questions.

One obvious point is that mutual regret and the multi-agent setting we have considered are only proxies for the sequential setting for transfer learning introduced in Section \ref{sec:vanilla}. It is an open problem whether one can achieve any results beyond mutual transfer that are able to quantify the benefit of samples collected when learning a specific task $M_\alpha$ for another particular task $M_{\alpha'}$. A~related question is that of obtaining bounds on the individual regret of a single agent in the considered multi-agent setting. Intuitively, if the mutual regret is smaller than the sum over the individual regret values of all agents, some agents have to profit from transfer. However, it seems to be  difficult to show that such an advantage exists for a particular agent.  

We have already pointed out that in principle any sensible RL algorithm with respective theoretical guarantees could be chosen instead of UCRL2 and be  modified to make use of additional samples. An obvious choice would be e.g.\ the algorithm proposed in \cite{boone} as it guarantees optimal regret  rates. One could even think of using samples from some algorithm to improve the performance of a different algorithm. In any case, it is an interesting open question whether some algorithms provably use additional samples collected during learning another task in a more efficient way than others. One can imagine that this is already determined by the regret bounds for the single-agent setting. However, this need not be the case in general.   

Last but not least, while regret is a natural and in our view the most competitive measure for RL, one could try to obtain similar results using e.g.\ sample complexity bounds instead.
In any of these directions there is a lot of room for future work and we hope that our research sparks some interest in providing more respective theoretical results for transfer in RL.

\appendix
\section{Proof of Theorem~\ref{thm}}\label{app}
First, note that the difference between the observed rewards $r^\alpha_t$ when agent $\alpha$ chooses action $a$ in state $s$ at step $t$ and the respective mean rewards $r^\alpha(s,a)$ is a martingale difference sequence and hence can be bounded by Azuma-Hoeffding (cf.~e.g. Lemma~10 in \cite{jaorau}), so that 
\[
    \sum_\alpha \sum_t \big(r^\alpha_t - r^\alpha(s_t^\alpha,a_t^\alpha) \big)  \leq \sqrt{\tfrac{5}{2}T\aleph \log \tfrac{8T\aleph}{\delta}}
\]
with probability at least $1-\big(\frac{\delta}{8\aleph T}\big)^{5/4} \geq 1-\frac{\delta}{12 (\aleph T)^{5/4}}$, similarly to eq.~(7) of~\cite{jaorau}. Accordingly, writing $v_k^\alpha(s,a)$ for the number of visits of agent $\alpha$ in $(s,a)$ during episode $k$ and defining $\Delta_k^\alpha := v_k^\alpha(s,a) (\rho^{*,\alpha} - r^\alpha(s,a))$ to be the regret of agent $\alpha$ in episode $k$, we can bound 
\begin{equation}\label{eqr1}
 R^{\rm m}_T = \sum_\alpha R^\alpha_T \,\leq\, \sum_\alpha\sum_k \Delta_k^\alpha + \sqrt{\tfrac{5}{2}\aleph T \log \tfrac{8 \aleph T}{\delta}}
\end{equation}
with probability $1- \frac{\delta}{12  (\aleph T)^{5/4}}$.

Analogously to App.~B.1 of \cite{jaorau} using our slightly adapted confidence intervals \eqref{eq:cr2} for the rewards, it can be shown that at each step $t$ the probability that there is an agent $\alpha$ whose true MDP $M^\alpha$ is not contained in its set of plausible MDPs $\mathcal{M}^\alpha(t)$ is bounded by $\tfrac{\delta}{15t^6}$. As shown in Sec.~4.2 of \cite{jaorau}, this implies that the respective regret caused by failing confidence intervals is bounded by
\begin{equation}\label{eqr2}
  \sum_\alpha\sum_k \Delta_k^\alpha \, \mathbb{1}\{M^\alpha \notin \mathcal{M}^\alpha_k\} \,\leq\, \aleph \sqrt{T}
\end{equation}
with probability $1- \frac{\delta}{12 (\aleph T)^{5/4}}$.

For the episodes $k$ in which $M^\alpha \in \mathcal{M}^\alpha_k$, one has analogously to eqs.~(14)--(18) of \cite{jaorau} that
\begin{align}
       \sum_\alpha\sum_k & \Delta_k^\alpha \, \mathbb{1}\{M^\alpha \notin \mathcal{M}^\alpha_k\}  \, \leq \\
       & \left( \sqrt{14 \log\left({\tfrac{2SA\aleph T}{\delta}}\right)} + 2 \right) 
              \sum_\alpha \sum_k \sum_{s,a} \frac{v_k^\alpha(s,a)}{\sqrt{\max{\{1,N_{t_k}^\alpha (s,a)\}}}} \nonumber \\
       & + D \sqrt{{14S\log\left({\tfrac{2A T}{\delta}}\right)}} \,  \sum_\alpha \sum_k \sum_{s,a} \frac{v_k^\alpha(s,a)}{\sqrt{\max{\{1,N_{t_k} (s,a)\}}}} \nonumber \\
       & +   \sum_\alpha \sum_k \sum_{t=t_k}^{t_{k+1}-1} \big( p(\cdot \mid s_t^\alpha,a_t^\alpha) - \mathbb{e}_{s_t^\alpha} \big) \mathbf{w}_k^\alpha, \label{eq:a1}
\end{align}
where $t_k$ denotes the initial time step of episode~$k$, $\mathbb{e}_i$ is the unit vector with $i$-th coordinate 1 and all other $S-1$ coordinates 0, and $\mathbf{w}_k^\alpha$ is a modified value vector (computed by extended value iteration for episode $k$, cf.\ footnote~\ref{fn:evi}) for agent $\alpha$ with $\|\mathbf{w}_k^\alpha\|_\infty \leq D/2$.

The last term of \eqref{eq:a1} is a martingale difference sequence and another application of Azuma-Hoeffding gives (similar to eq.~19 of \cite{jaorau} but now for a sequence of length $\aleph T$)
\begin{align}
     \sum_\alpha \sum_k \sum_{t=t_k}^{t_{k+1}-1} &\big( p(\cdot \mid s_t^\alpha,a_t^\alpha) - \mathbb{e}_{s_t^\alpha} \big) \mathbf{w}_k^\alpha   \,\leq\,  \nonumber \\
     &  D \sqrt{\tfrac{5}{2} \aleph T \log\big( \tfrac{8\aleph T}{\delta}}\big) + DSA\aleph \log_2\big(\tfrac{8T}{SA}\big)
\end{align}
with probability at least $1-\frac{\delta}{12 T^{5/4}}$, where the last term comes from a bound over the number of episodes, similar to Appendix B.2 of \cite{jaorau} (but now counted extra for each agent).

Finally, for the two triple sums in \eqref{eq:a1}, we have as shown in \cite{jaorau} that
\begin{equation}
    \sum_k \sum_{s,a} \frac{v_k^\alpha(s,a)}{\sqrt{\max{\{1,N_{t_k}^\alpha (s,a)\}}}}  \,\leq \, 
                \big(1+\sqrt{2}\big) \sum_{(s,a)} \sqrt{N_T^\alpha(s,a)}
\end{equation}
and similarly
\[
    \sum_\alpha \sum_k \sum_{s,a} \frac{v_k^\alpha(s,a)}{\sqrt{\max{\{1,N_{t_k} (s,a)\}}}}  \,\leq \, 
               \big(1+\sqrt{2}\big) \sum_{(s,a)} \sqrt{N_T(s,a)}.
\]
Observing that $\sum_{s,a} N_T^\alpha(s,a) = T$ for each $\alpha$ and $\sum_{s,a} N_T(s,a) = \aleph T$ it follows by Jensen's inequality that
\begin{equation}\label{eqr3a}
   \sum_\alpha \sum_k \sum_{s,a} \frac{v_k^\alpha(s,a)}{\sqrt{\max{\{1,N_{t_k}^\alpha (s,a)\}}}}  \,\leq \, 
                  \big(1+\sqrt{2}\big) \aleph \sqrt{SA T} .
\end{equation}
and
\begin{eqnarray}
   \sum_\alpha \sum_k \sum_{s,a} \frac{v_k^\alpha(s,a)}{\sqrt{\max{\{1,N_{t_k} (s,a)\}}}} &=&
     \sum_k \sum_{s,a} \frac{ \sum_\alpha v_k^\alpha(s,a)}{\sqrt{\max{\{1,N_{t_k} (s,a)\}}}}  \nonumber \\
     &\leq&  \big(1+\sqrt{2}\big)  \sqrt{SA\aleph T} .   \label{eqr3b}
\end{eqnarray}

In summary, we obtain from eqs. \eqref{eqr1}--\eqref{eqr3b} that the mutual regret is bounded by
\begin{align*}
  R^{\rm m}_T = \sum_\alpha R^\alpha_T  & \,\leq\, (D+1) \sqrt{\tfrac{5}{2}\aleph T \log \tfrac{8 \aleph T}{\delta}}  +  \aleph\sqrt{T}
    + DSA\aleph \log_2\big(\tfrac{8T}{SA}\big)\\
     & + \big(1+\sqrt{2}\big) \left(  \sqrt{{14\log\left({\tfrac{2A\aleph T}{\delta}}\right)}} + 2\right) \aleph \sqrt{SA T} \\
    & + \big(1+\sqrt{2}\big) D \sqrt{{14S\log\left({\tfrac{2A T}{\delta}}\right)}} \sqrt{SA\aleph T}    
\end{align*}
with probability $1-\frac{\delta}{4 T^{5/4}}$. Summing over all $T=2,\ldots$ shows that this bound holds simultaneously for all $T\geq 2$ with probability at least $1-\delta$.

It remains to simplify the bound. Summarizing terms, noting that $DSA\aleph \log_2{\left(\frac{8T}{SA}\right)} \leq \frac{2}{34}DS\sqrt{\aleph AT\log{\frac{\aleph T}{\delta}}}$ if  $T\leq 34^2A\aleph \log{\frac{\aleph T}{\delta}}$ (otherwise the theorem holds trivially, cf. App. B of \cite{jaorau}) we obtain
\begin{align*}
 R^{\rm m}_T  =  \sum_\alpha R^\alpha_T  & \,\leq\,  14 \aleph \sqrt{SA T \log\left({\tfrac{2A\aleph T}{\delta}}\right)} 
   + 15 DS  \sqrt{SA\aleph T \log\left({\tfrac{8A\aleph T}{\delta}}\right)} ,
\end{align*}
which completes the proof of Theorem~\ref{thm}.

\section{Proof of Theorem~\ref{thm2}}\label{app2}
The proof of Theorem~\ref{thm2} follows the same line as the proof of Theorem~\ref{thm} with a few minor adaptations due to the modified confidence intervals \eqref{eq:cr2} for the rewards, which however only causes slight changes in the constants. The main difference is that due to the modified confidence intervals, all terms linear in~$\aleph$ can be avoided:

First, all agents now share the same true MDP ${M}$ as well as the set of plausible MDPs $\mathcal{M}_k$, so that the probability that ${M}$ is not in $\mathcal{M}^\alpha_k$ neither depends on the single agents nor on their number $\aleph$. The respective regret over all agents then is bounded by $\sqrt{\aleph T}$ instead of $\aleph \sqrt{T}$ as in \eqref{eqr2}, however with the same error probability.

Further, using the modified confidence intervals instead of eq.~\eqref{eqr3a} one obtains
\begin{eqnarray*}
   \sum_\alpha \sum_k \sum_{s,a} \frac{v_k^\alpha(s,a)}{\sqrt{\max{\{1,N_{t_k} (s,a)\}}}}
         &=&  \sum_k \sum_{s,a} \frac{v_k(s,a)}{\sqrt{\max{\{1,N_{t_k} (s,a)\}}}} \\ [2\jot]
         &\leq&   \big(1+\sqrt{2}\big) \sqrt{SA \aleph T} ,   
\end{eqnarray*}
which also replaces the linear dependence on $\aleph$ by $\sqrt{\aleph}$ and allows to summarize terms similar to the proof of the original regret bounds of \cite{jaorau}. Details of the derivation are straightforward and are therefore skipped.

\subsection*{Acknowledgments}
We would like to thank the anonymous reviewers whose valuable comments greatly helped to improve the presentation of our results. The first author conducted work on of this paper during an internship for Universit\'{e} Paris-Saclay, ENS Paris-Saclay, France.

\subsection*{Declarations Concerning Funding and Competing interests}
No funding was received for conducting this study.
The authors have no relevant financial or non-financial interests to disclose.

\bibliography{RL}
\bibliographystyle{abbrv}

\end{document}